\documentclass[lettersize,journal]{IEEEtran}
\usepackage{biblatex} %Imports biblatex package
\usepackage{amsmath}
\usepackage{amssymb}
\usepackage{hyperref}
\usepackage{algorithm}
\usepackage{comment}

\usepackage{algorithmic}
\usepackage{booktabs}

\usepackage{bbm}

\usepackage{caption}  
\usepackage{subcaption}
\usepackage[nolist]{acronym}

\ifCLASSINFOpdf
  \usepackage[pdftex]{graphicx}
  \graphicspath{{../pdf/}{../jpeg/}}
  \DeclareGraphicsExtensions{.pdf,.jpeg,.png}
\else
  \usepackage[dvips]{graphicx}
  \graphicspath{{../eps/}}
  \DeclareGraphicsExtensions{.eps}
\fi

\usepackage{amsmath}

\usepackage{algorithmic}

\usepackage{array}
\usepackage{xcolor}

\newcommand{\orcid}[1]{\href{https://orcid.org/#1}{\textcolor[HTML]{A6CE39}{\aiOrcid}}}

\newcommand*{\vect}[1]{\boldsymbol{#1}}

\newcommand*{\bvect}[1]{\overline{\boldsymbol{#1}}}
\newcommand*{\tvect}[1]{\widetilde{\boldsymbol{#1}}}

\newcounter{remarknumber}
\begin{document}

\title{Enhancing UAV Search under Occlusion using Next Best View Planning}

\author{Sigrid Helene~Strand, Thomas Wiedemann, Bram Burczek, and Dmitriy Shutin,~\IEEEmembership{Senior Member,~IEEE}
        % <-this % stops a space
\thanks{This work was supported by the German Aerospace Center and the German Research Foundation (DFG) under Grant I 6046-N. (Corresponding author: Sigrid Helene Strand.)
 Sigrid Helene~Strand, Thomas Wiedemann, Bram Burczek, and Dmitriy Shutin are with the Swarm Exploration Group, Department of Communication Systems, Institute for Communications and Navigation, German Aerospace Center (Deutsches Zentrum für Luft- und Raumfahrt, DLR), Wessling 82234, Germany. (e-mail: sigrid.strand@dlr.de; thomas.wiedemann@dlr.de; bram.burczek@gmail.com; dmitriy.shutin@dlr.de). Thomas Wiedemann is also with Perception for Intelligent Systems (PercInS), TUM School of Computation, Information and Technology, Technische Universität München (TUM), 80992 München. (e-mail: thomas.wiedemann@dlr.de). }% <-this % stops a space
\thanks{Manuscript received 21 March 2025.}}
% The paper headers
\markboth{Journal of \LaTeX\ Class Files,~Vol.~14, No.~8, August~2021}%
{Shell \MakeLowercase{\textit{et al.}}: A Sample Article Using IEEEtran.cls for IEEE Journals}

%\IEEEpubid{0000--0000/00\$00.00~\copyright~2021 IEEE}
% Remember, if you use this you must call \IEEEpubidadjcol in the second
% column for its text to clear the IEEEpubid mark.

% make the title area
\maketitle

% As a general rule, do not put math, special symbols or citations
% in the abstract or keywords.

\begin{abstract}
    Search and rescue missions are often critical following sudden natural disasters or in high-risk environmental situations. The most challenging Search and Rescue missions involve difficult-to-access terrains, such as dense forests with high occlusion. Deploying unmanned aerial vehicles for exploration can significantly enhance search effectiveness, facilitate access to challenging environments, and reduce search time. However, in dense forests, the effectiveness of unmanned aerial vehicles depends on their ability to capture clear views of the ground, necessitating a robust search strategy to optimize camera positioning and perspective. This work presents an optimized planning strategy and an efficient algorithm for the Next Best View problem in occluded environments. Two novel optimization heuristics, a geometry heuristic, and a visibility heuristic, are proposed to enhance search performance by selecting optimal camera viewpoints. Comparative evaluations in both simulated and real-world settings reveal that the visibility heuristic  achieves greater performance, identifying over 90\% of hidden objects in simulated forests and offering 10\% better detection rates than the geometry heuristic. Additionally, real-world experiments demonstrate that the visibility heuristic heuristic provides better coverage under the canopy, highlighting its potential for improving search and rescue missions in occluded environments.

\end{abstract}

% Note that keywords are not normally used for peerreview papers.
\begin{IEEEkeywords}
Next Best View, Aerial Mapping, Unmanned Aerial Vehicle, Search and Rescue, Evolutionary Algorithms, Optimal Experimental Design
\end{IEEEkeywords}

% For peer review papers, you can put extra information on the cover
% page as needed:
% \ifCLASSOPTIONpeerreview
% \begin{center} \bfseries EDICS Category: 3-BBND \end{center}
% \fi
%
% For peerreview papers, this IEEEtran command inserts a page break and
% creates the second title. It will be ignored for other modes.
\IEEEpeerreviewmaketitle

%A forest with varied tree density and leaf coverage affects the ground visibility and ability to search for missing people. 

\section{Introduction}

\label{intro}

\IEEEPARstart{T}{he} equipment and methods currently used in \ac{SAR} operations are often inadequate for searches in obstructed or inaccessible terrain. Enhancing strategies for locating missing persons in such environments is therefore a critical objective. Difficult terrain poses significant obstacles to rescue missions, particularly when ground access is limited or impossible. In recent years, \ac{UAV}s have emerged as a valuable asset for \ac{SAR}, demonstrating remarkable effectiveness across various demanding landscapes and delivering promising results in urgent scenarios. However, certain terrains, such as dense forests, remain problematic for both human teams and \ac{UAV}s. Forest environments present unique challenges, including reduced visibility for sensors positioned above the canopy. Thermal cameras may struggle with heat reflectance from foliage during daylight hours and laser scanners often fail to differentiate human bodies on the forest floor. Therefore, we propose that strategically angled cameras can still provide useful visual data, improving search efficiency in these complex settings. 

In the literature, numerous path planning algorithms have revealed the possibilities of efficient search with \ac{UAV}s equipped with a camera. Several 3D exploration issues for aerial robotic platforms have been solved using a technique called \ac{NBV} \cite{Bircher2016RecedingH} \cite{Mandischer2022FindingMO}. The path planning algorithm can be adapted to solve various issues, such as 3D exploration of unknown space and \ac{SAR} operations with autonomous \ac{UAV}. Although the \ac{NBV} problem enables us to discover a camera angle that overcomes the issue of seeing in a dense forest, the question of what perspective is ideal remains unanswered.

Our approach, based on the \ac{NBV} scheme, attempts to improve \ac{SAR} by introducing unique heuristics to select the optimal view, which increases the chance of discovering missing persons. Furthermore, the developed heuristics are studied and evaluated according to their performance. In both simulated and real-world investigations, performance is assessed by the outcome of the NBV algorithm's amount of recognized hidden objects while applying the proposed heuristics.

\section{Related Work}
\subsection{Robotic Systems for Search and Rescue}
Scientists within the robotics community have been studying robots to acquire data for various purposes such as environmental monitoring, surveying, and rescue missions \cite{Dunbabin2012RobotsFE} \cite{https://doi.org/10.1002/rob.20222} \cite{app112210736}. Robotic systems have played a role in conducting rescue efforts as part of \ac{SAR} missions. In 2001, a single-robot system was first documented as being deployed in a disaster response. Since then, there have been many reported situations of single and multi-robotic deployment in response operations~\cite{Drew2021}. The prerequisites for using robots for this purpose are robustness, speed, and ease of use by human personnel. Even with cutting-edge robotics research, meeting this standard is often difficult \cite{https://doi.org/10.1002/rob.21887}. \ac{UGV}s have been utilized for numerous applications in the rescue response, such as searching for survivors \cite{1337826}, inspecting inside damaged buildings, and other hazardous areas following natural disasters \cite{10.7551/mitpress/9407.001.0001}. The main disadvantage is the difficulty of navigation in the complex surroundings \cite{https://doi.org/10.1002/rob.21887}. Rescue operations for \ac{UGV}s are limited to locations where the terrain is flat and the capacity to maneuver freely in space is considered limited.

Recent studies show interest in making \ac{SAR} missions more efficient by using \ac{UAV}s. They can navigate over terrain and have an overhead perspective, which is especially helpful in many rescue situations, such as dense map constructions for inspection \cite{Bircher2016RecedingH} and forest fire monitoring \cite{Yuan2015UAVbasedFF} \cite{https://doi.org/10.1002/rob.21887}.
Additionally, \ac{UAV}s are easy to develop, dynamic, and can be modified for specific tasks in \ac{SAR} applications \cite{drones6070154}. However, sensor payload is often restricted by the size and power of the \ac{UAV} \cite{https://doi.org/10.1002/rob.21887}.

There are three major motivations for performing \ac{SAR} research with UAVs.
First, research has shown that the exploration can be improved by making the \ac{UAV}s autonomous and developing an optimal path for the rescue scenario \cite{Wu2023AnAC}. 
Second, the volume of data acquired by the \ac{UAV} can be efficiently handled by applying a classification of objects of interest. 
This solution enhances the \ac{SAR} team's capacity to quickly locate objects of interest \cite{Aquino2023CNNCS}. 
Finally, establishing a communication framework among \ac{SAR} mission components can improve operations. In summary, \ac{UAV}s have proven to be valuable and play an important role in enhancing \ac{SAR} missions across various applications \cite{drones6070154}.

%\subsection{Unmanned Aerial Vehicles and Sensors}
\ac{UAV}s equipped with various sensors have been deployed for a variety of \ac{SAR} applications. \ac{RGB} cameras, thermal cameras, radars  and \ac{LiDAR} are some of the sensors that have been utilized to visualize surroundings or locate missing persons \cite{9664527}. A \ac{RGB} camera is often used on a \ac{UAV} due to its lightweight and ease of usage. The \ac{RGB} images can then be employed in conventional computer vision applications such as object recognition, monitoring, and surveillance. One downside is the limitation of light or occlusion. When there is no visibility, image quality and processing suffer significantly. Furthermore, thermal cameras have demonstrated advantages in the field due to their application for identifying heat sources. Thermal images have the same applications as \ac{RGB} photos but can capture images in low-light circumstances \cite{9664527} \cite{rs15184369}. However, when the sensor is used in environments with temperatures similar to those of persons, separating the person from the ground is challenging. Radars have proven beneficial in \ac{SAR} scenarios, such as locating survivors after earthquakes \cite{radar}. The sensor operates independently of light and weather conditions, but its resolution is lower compared to that of a \ac{LiDAR} sensor \cite{radar2}. \ac{LiDAR} sensors are frequently utilized to create a 3D model from the point cloud acquired. The \ac{LiDAR} sensor offers the advantage of functioning effectively in any lighting conditions while delivering valuable structural insights about the environment. However, it can struggle in adverse weather conditions, such as heavy rain or fog, which scatter the laser beams and reduce accuracy \cite{9664527}.

\subsection{The Next Best View Problem}
The \ac{NBV} problem has challenged scientists for many years \cite{Wong1999NextBV} \cite{4048302}. With solutions that can improve exploration, mapping and \ac{SAR} tasks, the \ac{NBV} plays a clear role in the field of robotics. The main challenge of the \ac{NBV} problem is determining the optimal view position and orientation for a camera or sensor installed on a robot. The optimal technique necessitates the definition of the ideal view; additionally, executing the task and navigating to the waypoints in the most efficient manner \cite{9134859}. The gain from solving such a problem is to more efficiently create a 3D object reconstruction \cite{Yan2023ActiveIO} and 3D exploration paths \cite{Bircher2016RecedingH}. Previous scientific work has shown various attempts to tackle the problem that was initially addressed in the 1980s by Connolly \cite{Connolly1985TheDO}. Iteratively selecting the optimal viewpoint to maximize the gathered information, ultimately achieves a comprehensive observation. Further, J. Mavier et al. describe an approach for obtaining 3D data of an unknown scene. The method focuses on occluded regions, which means that only the edges of the occlusions are modeled to determine the next step \cite{Maver1993OcclusionsAA}. 

More recent studies show how robotic platforms are used to autonomously explore unknown space based on the \ac{NBV} scheme. The methods, proposed by Scott et al. \cite{10.1145/641865.641868},  are divided into surface-based and volumetric approaches. In 2016, a volumetric receding horizon \ac{NBV} technique was introduced by Bircher et al. \cite{Bircher2016RecedingH}. For a computed random tree of a voxel-based area, the best branch is found, and its quality is defined by the quantity of undiscovered space available for exploration. The first edge of the best branch from the \ac{RRT} is always executed. A surface-based approach was suggested by Krainin et al. \cite{Krainin2011AutonomousGO}. The system is created for a robot to build a 3D surface model of an object based on a depth camera. Kriegel et al. \cite{Kriegel2015} focused on both of the approaches. A high-quality surface model is created from generated next-best-scans (NBS) and based on a selection criterion determined. 

\begin{comment}
    This work provides a surface-based technique for sampling \ac{NBV}s in partially known space that uses previous views' knowledge to calculate the next views. Similar to Bircher et al.'s \cite{Bircher2016RecedingH} voxel-based approach, the unexplored area is evaluated, but it is defined by the vertices of the surface model. Furthermore, this approach is compared to a developed heuristic based on the geometry of camera view constellations. 
\end{comment}

This work introduces a surface-based technique for sampling \ac{NBV}s in partially known space, utilizing prior views' knowledge to determine the next views. Similar to Bircher et al.'s \cite{Bircher2016RecedingH} voxel-based approach, the unexplored area is assessed, but here it is defined by the vertices of the surface model. Additionally, this approach is compared to a newly developed heuristic based on the geometry of camera view constellations, offering a novel method for efficiently planning the next viewpoints. Our contribution lies in the development of these two novel heuristics, which can be applied in \ac{SAR} missions by providing an effective strategy for selecting viewpoints, thereby enhancing mission efficiency. While we do not focus on person detection, our methods enable the identification of optimal viewpoints, which is a crucial step before detecting missing persons.

\section{Problem Description}
\label{sec:ProblemDescription}
%In \autoref{intro} the need of an optimised and efficient algorithm for search of missing people in forest is addressed. In this paper, an algorithm for an efficient path planning for identifying hidden objects is presented. 

For the scope of this paper, let us assume an outdoor environment cluttered with obstacles, like trees or bushes, that obstruct the view.
We assume that the scenery has been %observed and 
roughly mapped and reconstructed using photogrammetry methods from $\mathcal{C}_k=\{\vect{o}_k, \theta_k, \gamma_k\}$, $k=1,\ldots,K$, initial camera views.
Here each $k$th camera $\mathcal{C}_k$ is parameterized with a camera position $\vect{o}_k\in\mathbb{R}^3$ in 3D space, as well as camera orientation defined in terms of pitch $\theta_k$ and yaw $\gamma_k$ angles. 
\footnote{Note, that the roll angle of the camera is assumed to be fixed since it does not change the visibility of the scene substantially. 
Thus, only two angles are used to specify the orientation.}
The collection of the views is used to construct an initial, or \emph{a priori}, 3D model of the environment.
%We assume that the environment is \textit{a priori} given as a rough 3D model.
This initial model we represent by its surface as a triangular mesh $\mathcal{M}=\{\mathcal{V},\mathcal{F}\}$, where $\mathcal{V}=\{\vect{n}_1,\ldots,\vect{n}_N\}$ is a set of vertices with coordinates $\vect{n}_i \in \mathbb{R}^3$, $i=1,\ldots,N$, and $\mathcal{F}$ is a set of corresponding triangular faces.
%, with each face being a plane in 3D defined on neighboring vertices.

%We assume that the environment was reconstructed by photogrammetry methods based on multiple $K$ cameras.
%Each camera is characterized with a specific view $C_k$, $k=1,\ldots,K$, which includes camera position $p_k$ and rotation $R$ in 3D. 
%Initial camera views $\mathcal{C}[0]=\mathcal{C}_{\text{init}}$ at the moment of time $t=0$ will thus represent an initial state of the algorithm.
%As the algorithm progresses, $\mathcal{C}[t]$ will vary, reflecting both changing camera orientations as well as (possibly) their number. 

Within the scenery, an unknown number of objects of interest, e.g. missing people or damaged structures, are located at unknown positions.
Due to the obstacles and obstructed view, these objects might not be fully visible in the initial top-down views $\mathcal{C}_k$, $k=1,\ldots,K$; instead, they can be obstructed by foliage or other obstacles within the cameras field of views.
Our goal is to iteratively calculate a \emph{\ac{NBV}} configuration $\mathcal{C}_{K+1}$ that includes new positions $\vect{o}_{K+1}$ and orientations $(\theta_{K+1},\gamma_{K+1})$ of a camera carried by a \ac{UAV}, which would increases the visibility of  hidden objects in one of the camera's views.  
 
%that increases our chance to see one of the hidden objects in one of the camera's images.  

%that all objects of interest are considered with an unknown position and unknown amount. 
%Some objects may be seen from initial vertical camera positions, however the amount of hidden objects are unknown. A need of calculating the initial position and next view where an potential object may be seen present. Additionally, the planned path of views need to be considered the most efficient path. 

Finding the \ac{NBV} can be formulated as an optimization problem.
However, since the location of the hidden objects is generally unknown, these cannot be  included directly in the optimization objective.
Instead, we propose and investigate optimization heuristics that can be used to score different views in the optimization routine.
%The \ac{NBV} algorithm consider several heuristics that each indicates the weight of the \ac{NBV}. 
%How successful these heuristics work in finding the \ac{NBV} needs to be compared and evaluated in both simulated and real experiments. 
%For a set of calculated camera positions from the heuristics the best solution to the problem is to find the most amount of missing objects. In a perfect camera constellation, the total amount of people would be found, but in this project the best solution is to find as many as possible.
Our goal is thus to place cameras optimally so that they can better ``illuminate'' hidden objects in the scene.
%The actual location of the hidden objects is assumed \emph{a priori} unknown.
%Therefore, we cannot formulate an optimization criterion based on the hidden object directly; instead, we need to rely on an indirect approach. 
We propose two heuristics that, when optimized, indirectly improve the view of the scene and thus increase the chance of finding hidden objects.
%In this article, we propose and analyze two different optimization heuristics for \ac{NBV} planning.
%The heuristics are used as fitness functions in the view optimization explained later in this chapter.
This procedure is detailed in the following.

%which are not of concern in this article.
%The where all calculated \acp{NBV} are represented by the set $\mathcal{C}_{\text{NBV}}$.
%The set of all camera views is $\mathcal{C}=\mathcal{C}_{\text{init}} \bigcup\mathcal{C}_{\text{NBV}}$.
%Our \ac{NBV} planning algorithm is an iterative approach that successively plans new camera views.
%So overtime, the set of camera $\mathcal{C}_{\text{NBV}}$ and $\mathcal{C}$ grow. 
%To indicate the set at a particular iteration $k$ of our approach, we denote $\mathcal{C}[k]$.

\section{Next Best View Optimization Approach}
When formulated as an optimization problem, finding the \acf{NBV} requires an appropriate definition of a fitness (objective) function.
In this work, we introduce two distinct heuristics to serve as fitness functions for the \ac{NBV} optimization problem. Our goal is to develop an effective yet straightforward heuristic that indirectly enhances object visibility.

\subsection{Visibility Fitness}
\label{sec:visibility_fitness}
The motivation for the first proposed heuristic is to maximize the number of vertices in $\mathcal{V}$ that are seen by a new camera view $\mathcal{C}_{K+1}$ by changing its position and orientation.
Moreover, the heuristic should reward a new camera view, if a vertex is not readily seen by the other, already placed cameras $\mathcal{C}_k$, $k=1,\ldots,K$.
At the same time, we penalize the objective if too many cameras observe the same vertex, thus not delivering any additional information.

Let us calculate the visibility of all vertices in $\mathcal{V}$ by a new view $\mathcal{C}_{K+1}$.
We represent the visibility as an $N$-dimensional binary vector $\vect{w}\in \{0,1\}^{|\mathcal{V}|}$.
Each element $w_i\triangleq [\vect{w}]_i$, $i=1,\ldots,N$, encodes if a vertex $i$ is visible by the camera $\mathcal{C}_{K+1}$.
In particular,
\begin{equation}
\begin{split}
\label{eq:visibility_matrix}
w_{i}=\left\{\begin{array}{l}
1; \text { if $\vect{n}_i$ is visible by $\mathcal{C}_{K+1}$}\\
0; \text { else }
\end{array}\right.,
\end{split}\quad \forall \vect{n_i} \in \mathcal{V}.
\end{equation}
To check for visibility we use a simple ray tracer.
Namely, if the ray emitted from the camera position $\vect{o}_{K+1}$ to the vertex $\vect{n}_i\in\mathcal{V}$ intersects with any face in $\mathcal{F}$, we consider the vertex as not visible.
The same goes for the case where the vertex $\vect{n}_i\in\mathcal{V}$ is outside the field of view of a new camera $\mathcal{C}_{K+1}$.

In the same way, we define the visibility matrix for already available camera views $\vect{M}\in \{0,1\}^{|\mathcal{V}|\times K}$, with $i-k$th element $[\mathbf{M}]_{i,k}$ given as:
\begin{equation}
\begin{split}
\label{eq:visibility_matrix_1}
    [\mathbf{M}]_{i,k}&=\left\{\begin{array}{l}
1; \text { if $\vect{n}_i$ is visible by $\mathcal{C}_k$}\\
0; \text { else }
\end{array}\right.,\\
\forall \vect{n}_i& \in \mathcal{V},\quad k=1,\ldots,K.
\end{split}
\end{equation}
%In other words, one can consider the matrix $\vect{M}$ as a table encoding if a vertex $i$ is seen by a camera $k$.
For each vertex, we can calculate the number of cameras that can see the respective vertex as:
\begin{equation}
    m_i = \sum_{k=1}^K [\mathbf{M}]_{i,k},
    %= \mathbf{M}\mathbbm{1},
\end{equation}
which is an $i$th row sum of the matrix $\vect{M}$.
%where $\mathbbm{1}=[1,1,\ldots,1]$ is a $K$-dimensional vector of ones.  
We now define the \emph{Visibility Fitness} function as:
\begin{equation}
    J_v(\mathcal{C}_{K+1}) = \sum_{i=1}^{N} \alpha_i w_{i}
    %J_v(\mathcal{C}_{K+1}) = \sum_{i=1\forall \vect{x_i} \in \mathcal{V}} \alpha_j w_{j}
    \label{equ:VisibilityFitness}
\end{equation}
where the weights $\alpha_i$, $i=1,\ldots,N$, are designed as  
\begin{equation}
\alpha_i = \frac{1-\tanh\left(m_i -  3\right)}{2}.
\label{equ:scalin_fucntion}
\end{equation}

\begin{figure}[t]
\centering
\centering
 \includegraphics[trim={0cm 0cm 0cm 0.0cm},clip,width=0.85\linewidth]{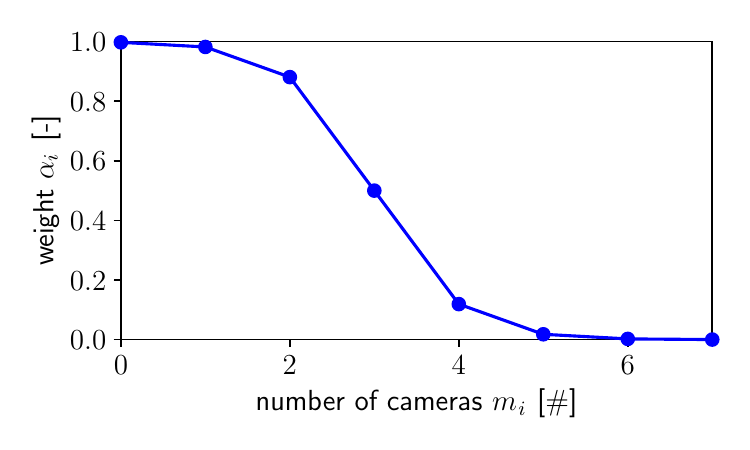}
\caption{The scaling function for the weights $\alpha_i$ in \eqref{equ:scalin_fucntion} versus vertex visibility.
}
\label{fig:weights}
\end{figure}

It can be seen that the weights $\alpha_i$, $i=1,\ldots,N$, are selected to reward vertices with low visibility $m_i$, while at the same time scaling down vertices that are already visible from multiple cameras.
%In other words, we rewards if many entries in $\vect{w}$ are not zero.
%That means if many vertices are visible by the new camera.
%At the same time, the weights $\alpha_j$ reduce the reward if a vertex is already visible by a lot of other cameras.
The scaling function \eqref{equ:scalin_fucntion} is visualized in Figure~\ref{fig:weights} to illustrate this behavior.

Finally, placing a new camera can be posed as a problem of maximizing the Visibility Fitness $J_v(\mathcal{C}_{K+1})$
\begin{align}
    \label{eq:visibility_heuristic}
      \max_{\vect{o}_{K+1},\theta_{K+1},\gamma_{K+1}} 
  J_v(\mathcal{C}_{K+1})
\end{align}
%\begin{align}
%    \label{eq:visibility_heuristic}
%      \max_{\vect{o}_{K+1},\tiny\begin{bmatrix}
%       \theta_{K+1},\\
%       \gamma_{K+1}
%  \end{bmatrix}} 
%  J_v(\mathcal{C}_{K+1})= 
%  \max_{\vect{o}_{K+1},\tiny\begin{bmatrix}
%       \theta_{K+1},\\
%       \gamma_{K+1}
%  \end{bmatrix}}\,
%  \sum_{i=1}^N \sum_{i=1}^{K} \alpha_i w_{i}
%\end{align}
where the optimization is performed with respect to the parameters $\vect{o}_{K+1}$, $\theta_{K+1}$ and $\gamma_{K+1}$ of the next $K+1$th camera.

%The coverage fitness heuristic is based on the amount of vertices seen in the environment by each camera. The relation between the visible  $vertex_m$ seen in $camera_n$ creates the visibility matrix, $V_E$, where $V_E$ consist of for all cameras and vertices in the environment, $E$. The relation is given in \autoref{eq:visibility_matrix}. 

%If a position $m,n$ in $V_E$ is given 1, the $camera_n$ have observed the $vertex_m$, and the vertex is therefore visible. When finding the amount of cameras that can see the $vertex_m$, the rows can simply be added up as in \autoref{eq:visibility_camera}. Here, the $V_C$ is the vector for the number of cameras where $vertex_m$ is visible.

%\begin{equation}
%\label{eq:visibility_camera}
%\underset{m, 1}{\mathbf{V_C}}=\overset{n}{ \underset{m=1}%{\sum}} \hspace{0.1cm} V_{E}
%\end{equation}

%Furthermore, the weights of visibility can be calculated for each $vertex_m$ in \autoref{eq:weights}. 

%\begin{equation}
%\label{eq:weights}
%\underset{m, 1}{\mathbf{P}} =  \frac{1}{2} -  \frac{1}{2} \cdot \tanh\left(\underset{m, 1}{\mathbf{V_C}} -  3.\right)
%\end{equation}

%Conclusively, the vertex with highest priority of the vertices, $\underset{m, 1}{\mathbf{P}}$, from \autoref{eq:weights} can be found by finding the maximum value. The heuristic is updated when a new camera is added for each iteration.
%\\

\subsection{Geometry Fitness}\label{sec:geometry_fitness}
Another criterion that we propose is based on geometrical properties.    
%Specifically, we consider the projection of vertexes $\vect{x_i} \in \mathcal{V}$, $i=1,\ldots,N$, from a 3D space to our the 2D camera images, which can be approximated with a linear transformation.
%In general, this projection incurs a nonlinear transformation, that can be locally approximated with a linear one. 
%The curvature of the corresponding transformation at the linearization point can be used to quantify the  quality, or rather precision, of the operation.
Specifically, we consider the information of the geometric constellation between the camera views to find the next best suitable camera $\mathcal{C}_{K+1}$, while accounting for the geometry of already placed cameras $\mathcal{C}_k$, $k=1,\ldots,K$. 
This geometry plays a role in the photogrammetric reconstruction of an individual vertex $\vect{n}_i\in\mathcal{V}$, $i=1,\ldots,N$ \cite{camera}.
Indeed, the computation of the vertex $\vect{n}_i$ using cameras $\mathcal{C}_k$, $k=1,\ldots,K$ can be interpreted as (in general) a nonlinear transformation between multiple 2D image spaces into a global 3D space. 
The curvature of this transformation in the vicinity of vertex coordinates in the object space can be interpreted as a precision of the transformation.
Alternatively, the inverse curvature can be interpreted as a (local) confidence, or uncertainty, ellipsoid of the transformation between image space and the global coordinates of the vertex.
The volume of this ellipsoid, as described by the determinant of the resulting inverse curvature matrix, can be naturally used as a criterion for camera placing: new camera position $\mathcal{C}_{K+1}$ is selected to reduce the resulting uncertainty ellipsoid.
Let us discuss these steps now in a bit more detail.
To this end, we first have a closer look at the relation between image and object space and a corresponding linearization.

\subsubsection{Relation between image and object space}
To understand the relation between the coordinates in the camera's image space and the coordinates in global space we need to define a coordinate transformation.
As a first step, we define the transformation of the \emph{global} 3D space into a \emph{local} camera 3D frame.
The local camera frame is defined with respect to camera orientation, such that its origin is located in the camera's focal point and the positive direction of the z-axis is alight with the direction of view.
The corresponding transformation specifically for the $k$th camera can be expressed as
\begin{equation}
\label{eq:correspondence}
    %\boldsymbol{x} = \boldsymbol{x}_{C_k} + \boldsymbol{R}_{C_k} \cdot \boldsymbol{x'},
    \vect{n} = \vect{o}_{k} + \vect{R}_{k}(\theta_k,\gamma_k)\vect{n'},
\end{equation}
where $\vect{n}$ and $\vect{n}'$ are vertices in the global and local coordinate frames, respectively, and $\vect{R}_{k}(\theta_k,\gamma_k)$ is (orthogonal) rotation matrix defined by the orientation angles $\theta_k$ and $\gamma_k$ of the corresponding camera $\mathcal{C}_k$.
Note that we implicitly assume that the local coordinate system is a function of a particular camera; as such, we generally have $K$ different local coordinate frames, and a single global one.   

From \eqref{eq:correspondence} is becomes obvious that 
%By inverting \autoref{eq:correspondence} we get:
\begin{equation}
\label{eq:inverted}
\vect{n'} = \vect{R}_{k}(\theta_k,\gamma_k)^T(\vect{n}-\vect{o}_{k}),
\end{equation}
where we used the fact that for orthogonal matrices $\vect{R}^{-1}=\vect{R}^T$.
The expression \eqref{eq:inverted} provides a mapping from the global (or object) coordinates $\vect{n}$ into the local camera frame $\vect{n}'$. 
Now, assuming that we have a non-distorted image with vertex coordinates $\vect{n'}=[x',y',z']^T$ and a camera modeled as a pin-hole camera with a focal length $f$, we can express 2D pixel coordinates $\vect{p}(\vect{n})=[u(\vect{n}),v(\vect{n})]^T$ of a vertex $\vect{n}$ as:
\begin{equation}
\label{eq:func_x_y}
    u(\vect{n}) =f \frac{x'}{z'},\quad
    v(\vect{n}) =f \frac{y'}{z'}.%\label{eq:func_y}
\end{equation}
The latter transformation expressions are also known as co-linearity equations \cite{luhmann}.

\subsubsection{Linearization of the co-linearity equation}

Inspecting \eqref{eq:func_x_y} reveals that the transformation between an arbitrary point $\vect{n}$ in a global coordinate system into pixel coordinates $(u,v)$ is nonlinear.
Consider now the $k$th camera $\mathcal{C}_k$; furthermore, let us consider a first-order Taylor expansion of this mapping around a mesh vertex $\vect{n}_i\in\mathcal{V}$ for the $k$th camera: 

%So far for the transformation we considered a single vertex and a single camera.
%From now on, we use the notation $\boldsymbol{u}_k(\boldsymbol{x}_i)=[u_k(\boldsymbol{x}_i),v_{k}(\boldsymbol{x}_i)]^T$ to indicate the transformation of vertex $\vect{x_i} \in \mathcal{V}$ to the image space of camera $C_k$ with $k=1,\ldots,K$.
%The transformation is non-linear as can be seen from \autoref{eq:func_x} and \autoref{eq:func_y}.
%However, the transformation can be linearized by a Taylor expansion as
%It is straightforward to conclude that
\begin{equation}
 \vect{p}_k(\vect{n}) \approx \vect{A}_{k,i}\vect{n} + \vect{c}_{k,i},
\end{equation}
where $\vect{c}_{k,i}$ is some constant term, and $\vect{A}_{k,i}$ is a Jacobian of the co-linearity equations defined as 
\begin{equation}
    \label{equ:colinearity_design}
\vect{A}_{k,i}=
%\left[\begin{array}{lll}
%\frac{d u_k(\boldsymbol{x})}{dx}\big|_{\boldsymbol{x}_i} &
%\frac{d u_k(\boldsymbol{x})}{dy}\big|_{\boldsymbol{x}_i} &
%\frac{d u_k(\boldsymbol{x})}{dz}\big|_{\boldsymbol{x}_i}\\
%\frac{d v_k(\boldsymbol{x})}{dx}\big|_{\boldsymbol{x}_i} &
%\frac{d v_k(\boldsymbol{x})}{dy}\big|_{\boldsymbol{x}_i} &
%\frac{d v_k(\boldsymbol{x})}{dz}\big|_{\boldsymbol{x}_i}\\
%\end{array}\right],
    \left.\begin{bmatrix}
        \nabla u(\vect{n})^T \\
        \nabla v(\vect{n})^T \\
    \end{bmatrix}\right|_{\vect{n}=\vect{n}_i},
\end{equation}

The matrix $\vect{A}_{k,i}$ thus represents a projection of a single vertex to corresponding image coordinates. 
%However, we can stack design matrices to project a single vertex to all camera images at once:
For all available cameras $k=1,...,K$, we can equivalently write 
\begin{equation}
\label{eq:linearization1}
\underbrace{
\begin{bmatrix}
    \vect{p}_1(\vect{n}) \\
    \vdots\\
    \vect{p}_K(\vect{n}) 
\end{bmatrix}}_{\triangleq\tvect{p}(\vect{n})} \approx 
\underbrace{
\begin{bmatrix}
    \vect{A}_{1,i} \\
    \vdots\\
    \vect{A}_{K,i} 
\end{bmatrix}}_{\triangleq\tvect{A}_{i}}  \vect{n} + \tvect{c}_{i},
\end{equation}
which summarizes coordinates of the \emph{same} vertex $\vect{n}$ as seen by different cameras in the corresponding $K$ camera images.

For the all available vertices $\vect{n}_i\in\mathcal{V}$ we can generalize \eqref{eq:linearization1} as follows
%Furthermore, we can write the projection of all vertices to all camera images by:
\begin{equation}
\label{eq:linearization2}
    \begin{bmatrix}
        \tvect{p}(\vect{n}_1)\\
        \tvect{p}(\vect{n}_2)\\
        \vdots\\
        \tvect{p}(\vect{n}_N)
    \end{bmatrix}\approx 
    \underbrace{
    \begin{bmatrix}
        \tvect{A}_1 & & &\\
        & \tvect{A}_2  & &\\
        && \ddots&\\
        &&& \tvect{A}_N\\
    \end{bmatrix}}_{\bvect{A}}  
    \begin{bmatrix}
        \vect{n}_1\\
        \vect{n}_2\\
        \vdots\\
        \vect{n}_N
    \end{bmatrix} + \mathrm{const},
\end{equation}
which is essentially an approximate affine mapping of the vertices $\mathcal{V}$ onto the pixel coordinates of the images for all $K$ cameras.

In practice, we are much more interested in inverting the mapping \eqref{eq:linearization2} to compute coordinates $\vect{n}$ from the corresponding images, i.e., compute from pixel coordinates the corresponding vertex coordinates in a global coordinate frame.
The accuracy of this inverse mapping based on \eqref{eq:linearization2} can be well measured with the dispersion matrix $\vect{Q}$ given as
\begin{equation}
    \label{eq:covariance_matrix}
    \vect{Q}= \left( \bvect{A}^T \bvect{A} \right)^{-1}=
    \begin{bmatrix}
        (\tvect{A}_1^T\tvect{A}_1)^{-1} & &\\
        & \ddots&\\
        && (\tvect{A}_N^T\tvect{A}_N)^{-1}\\
    \end{bmatrix}    
\end{equation}
Note that $\vect{Q}$ is computed based on images from $K$ cameras and thus provides certainty information about coordinates in a global coordinate space based on the available data.
However, thanks to the mapping \eqref{eq:linearization2}, we can now compute the impact of adding a new $K+1$ camera on the corresponding matrix $\vect{Q}_{K+1}$.

Let $\vect{o}_{K+1}$ and $\theta_{K+1},\gamma_{K+1}$ denote some arbitrary position and orientation of yet-to-be-placed camera $\mathcal{C}_{K+1}$.
From \eqref{equ:colinearity_design} we can compute the corresponding Jacobian $\vect{A}_{K+1,i}$ as if we have a camera placed at $\vect{o}_{K+1}$ and oriented it according to angles $(\theta_{K+1},\gamma_{K+1})$.
With the new camera, the matrices $\tvect{A}_{i}$ in \eqref{eq:linearization1} and the corresponding products $\tvect{A}_{i}^T\tvect{A}_{i}$, $i=1,\ldots,N$, in \eqref{eq:covariance_matrix} are extended as follows
\begin{align}
    &\tvect{A}_{i}(\mathcal{C}_{K+1})\triangleq 
    \begin{bmatrix}
     \tvect{A}_{i}\\
     \vect{A}_{K+1,i}
    \end{bmatrix},\\
    \tvect{A}_{i}(\mathcal{C}_{K+1})^T&\tvect{A}_{i}(\mathcal{C}_{K+1})\triangleq 
     \sum_{k=1}^{K+1}\left(\vect{A}_{k,i}^T\,\vect{A}_{k,i}\right).\nonumber
\end{align}
Note that we indicate an explicit dependency of $\tvect{A}_{i}(\mathcal{C}_{K+1})$ on the location and orientation of the new camera.
This allows us to compute the corresponding would-be dispersion matrix $\vect{Q}(\mathcal{C}_{K+1})$ simply by including an additional contribution from the new camera.

\subsubsection{Optimality criterion}
Now we can measure how the total dispersion changes due to placing a new camera $\mathcal{C}_{K+1}$.
%We have now found the relation between the image and the global space. 
%Additionally, we have calculated the covariance matrix of the projection of our vertices $\boldsymbol{x}_i \in \mathcal{V}$ to all images of the cameras $C_k\quad k=1,\ldots,K$.
This can be achieved using classical criteria from optimal experiment design theory \cite{fedorov1972theory}.
In particular, we will use a so-called D-optimality criterion, which measures the volume of the resulting uncertainty ellipsoid defined by the covariance matrix $\vect{Q}(\mathcal{C}_{K+1})$ as given by its determinant.
A smaller volume of the uncertainty ellipsoid obtained by adding a new camera $\mathcal{C}_{K+1}$ implies a more precise reconstruction of the 3D points in the global coordinates from individual camera images.        
Let us now define the corresponding NBV cost function with respect to a new camera $\mathcal{C}_{K+1}$ as
\begin{equation}
\begin{split}
  J_d(\mathcal{C}_{K+1})= det \left( \vect{Q}(\mathcal{C}_{K+1})^{-1}\right) =\\=\prod_{i=1}^N det  \left( \tvect{A}_{i}(\mathcal{C}_{K+1})^T \tvect{A}_{i}(\mathcal{C}_{K+1}) \right)=\\=\prod_{i=1}^N det  \left( \sum_k^{K+1} \vect{A}_{k,i}^T \vect{A}_{k,i}\right)
  \label{eq:determinant}
  \end{split}
\end{equation}

where we exploit the facts  $\vect{Q}(\mathcal{C}_{K+1})$ is a diagonal block matrix.
Note that we use an inverse of $\vect{Q}(\mathcal{C}_{K+1})$, which is equivalent to an information matrix.
As a result, maximizing $J_d(\mathcal{C}_{K+1})$ with respect to the parameter $\vect{o}_{K+1}$ and $\theta_{K+1},\gamma_{K+1}$ of the $K+1$th camera is equivalent to increasing the information about reconstructed 3D points in global coordinate system.  
Now, the placement of the new camera can be posed as the following optimization problem:
\begin{align}
    \label{eq:geo_heuristic}
      \max_{\vect{o}_{K+1}, \theta_{K+1},\gamma_{K+1} }
  J_d(\mathcal{C}_{K+1})
\end{align}
where the optimization is performed over the the position $\vect{o}_{K+1}$ and orientation $\theta_{K+1},\gamma_{K+1}]$ of the new $K+1$ camera.

Let us stress that although \eqref {eq:geo_heuristic} does require computing the determinant for each potential camera position, individual terms $\vect{A}_{k,i}$ as defined in \eqref{equ:colinearity_design} required to evaluate it, are only $3\times 3$ matrices. 
The corresponding determinant can be easily computed  in closed form and is thus easy to evaluate for each vertex $i=1,\ldots,N$ in the mesh $\mathcal{M}$.

\begin{figure*}[ht] % Use figure* for a figure spanning both columns 
\centering \includegraphics[width=0.8\textwidth]{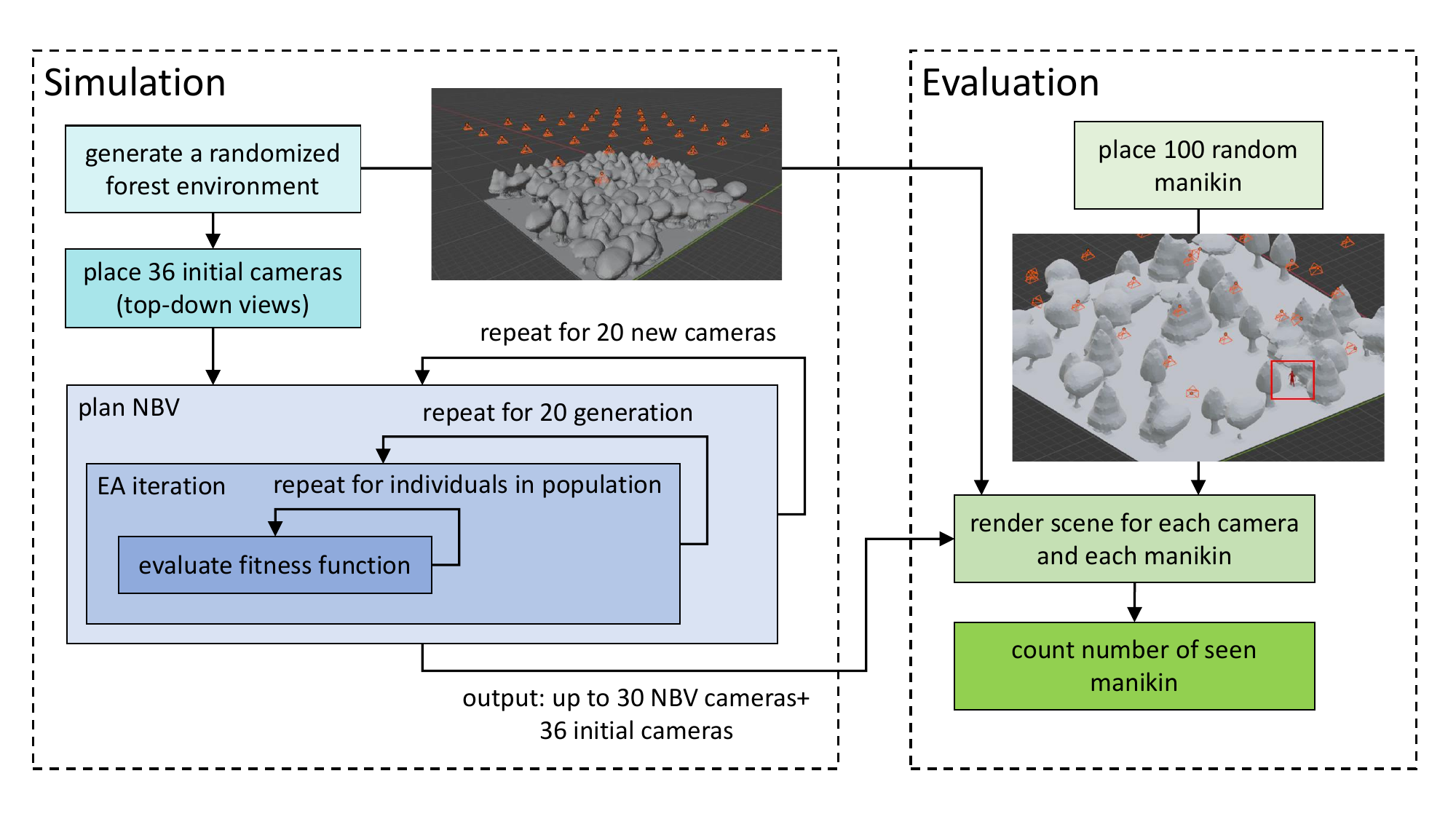} % Adjust the width parameter as needed 
\caption{Workfow of simulation experiments to evaluate the proposed \ac{NBV} planning.
} \label{fig:workflow_sim} \end{figure*}

\subsection{Optimization algorithm}
In the search of the \ac{NBV}, our goal is to maximize the fitness functions \eqref{equ:VisibilityFitness} or \eqref{eq:geo_heuristic} by placing new cameras $\mathcal{C}_{K+1}$ in an optimal way. 
While the computation of the fitness function for each individual camera position can be quite moderate, the search through the parameter space $\{\vect{o}_{K+1},\theta_{K+1},\gamma_{K+1}\}$ to maximize the corresponding fitness functions can be quite prohibitive.
Note that brute-force optimization of the constructed objective functions is impractical, as well as, gradient-based optimization algorithms.
%Also, the computation of the gradient is non-trivial; 
Both \eqref{eq:visibility_heuristic} and \eqref{eq:geo_heuristic} are non-convex, having multiple local optima, which renders the use of gradient-based approaches inappropriate.
%In search of the \ac{NBV}, the heuristics presented in the previous section will be utilized to find a good next view, i.e. the position and orientation of a camera $\mathcal{C}_{K+1}$.
%Essentially, the heuristics score the given 3D scene, containing the model %$\mathcal{M}$ and cameras $C_k, k=1...K+1$ according to one of the fitness functions presented before.
%The goal is to maximize the score by placing the new camera $\mathcal{C}_{K+1}$ in an optimal way.
%Let us now consider a practical approach to solve optimizations \eqref{eq:visibility_heuristic} and \eqref{eq:geo_heuristic} efficiently.

%This creates a setup for the following optimization problem:
%\begin{equation}
%\max_{\vect{x}_{\mathcal{C}_{K+1}},\vect{R}_{\mathcal{C}_{K+1}}} J_*,
%\end{equation}
%where $J_*$ is either chosen as $J_v$ or $J_d$.
%Considering a 3D environment with occluded areas and the already placed cameras $C_k, k=1...K$, the optimization is non-convex in general.
%This makes the optimization problem very difficult to determine the "best" camera view.
%For example, methods like gradient descent cannot be applied, since most likely they will get stuck in a local minimum or plateaus of the non-convex fitness function and reach no global optimum.
So, we propose to utilize an \textit{\ac{EA}} to address the considered optimization problems~\cite{EA1} \cite{DEAP_JMLR2012}.
In this way, we can cope with the high number of local optima in our optimization objective. 
Note that while \ac{EA} is a stochastic optimization approach, in general, it does not guarantee to reach a globally optimal solution in a finite time.
However, it might nonetheless be ``sufficiently'' close to the optimum, ensuring a good \emph{practical} solution rather than the best possible one.
%For that reason, we chose an \textit{\ac{EA}} to solve the optimization problem.
%In this way, we can cope with the high number of local minima in our optimization objective.
%However, at the same time as a stochastic optimization approach, we also cannot guarantee to reach a globally optimal solution in finite time.
%We are willing to take this compromise, as a global maximum will not contribute significantly more to our research goal compared to a solution close enough to the optimum.

In our case for the \ac{EA} algorithm we define an ``individual'' as  a particular combination of parameters $\vect{o}_{K+1}$, $\theta_{K+1}$, and $\gamma_{K+1}$ of a potential next camera view $\mathcal{C}_{K+1}$.

%\textit{individual} in our \ac{EA} optimization is the combination of parameters $\vect{x}_{\mathcal{C}_{K+1}}$, $\vect{R}_{\mathcal{C}_{K+1}}$ of a potential next camera view $\mathcal{C}_{K+1}$.
We will restrict the parameter space of possible camera positions to certain values.
This is done not only to reduce computational complexity but also to account for the physical constraints of a 3D scenery.  
First, we bound the z-component of $\vect{o}_{K+1}$ \--- the camera height \---  to a certain range above ground to avoid collision with the scene.
Also, X- and Y-component of $\vect{o}_{K+1}$ are restricted to a range close to the area of interest since camera locations that are far away from the scene could be excluded immediately as sub-optimal.
%Depending on the 3D scene, we restrict the range of the parameter space to certain values.
%E.g., we restrict the height (z-component of $\vect{x}_{\mathcal{C}_{K+1}}$) to a certain height over the ground to avoid collision with the scene.
%Further, in the orientation we fix the roll angle of the camera since it does not change the visibility of the scene substantially.
%Also x- and y-component of $\vect{x}_{\mathcal{C}_{K+1}}$ are restricted to be close to the area of interest since a camera positioned kilometers away can be excluded right from the beginning as sub-optimal

A population of a single generation in our \ac{EA} optimization includes $I$ such individuals.
Each individual in the population is initialized with random values for the first generation, yet adhering to the constraints of the search space, as defined above.
We then use either the Visibility Fitness $J_v$ or Geometry Fitness $J_d$ to determine the performance of each individual in the population, i.e., how well the camera is placed according to the proposed heuristic.
Depending on the fitness, the best individuals from the population are selected for the reproduction of the next generation.
Therefore, we select three individuals randomly from the old population and the best one is selected.
This process is repeated until the new generation is fully populated.
Selected individuals are then consecutively paired and recombined to produce new "offspring" individuals, a process known as crossover~\cite{EA1}.
Crossover allows the algorithm to explore new parts of the search space by combining different characteristics (genes) of the parent solutions. 
In our case, we make use of a two-point crossover.
The process involves selecting two points (or positions) in the parent genetic representation (here camera parameters) and swapping the genetic material between these points to generate a new individual.
In addition, a small portion of the offspring is randomly altered, which in an \ac{EA} algorithm is known as mutation. 
We use Gaussian mutation, where a gene, i.e. camera parameter, in an individual’s genome, is altered by a small randomly chosen value from a Gaussian distribution with zero mean and variance one. 
While mutation also allows the algorithm to explore new parts of the search space by making random changes, by making small changes mutation primarily helps to fine-tune existing good solutions.
The overall cycle is then repeated for multiple generations.

%The population of one generation in our \ac{EA} optimization consists of several such individuals.
%Each individual in the population is initialized with random values for the first generation.
%One of the previously defined fitness functions is then used to determine if an individual is good or bad, i.e., how well the camera is placed according to the heuristic.
%Depending on the fitness, the fittest individuals are selected for reproducing the next generation.
%Selected individuals are paired and recombined to produce new "offspring" individuals (crossover).
%In addition, a small portion of the offspring is randomly altered (mutation).
%This leads to variation in the population and helps to explore new parts of the solution space.
%The new generation of individuals then completely or partially replaces the old generation; True to the motto: "survival of the fittest". 
In our implementation, we make use of the DEAP python library \cite{EA1}.
The performance of the \ac{EA} algorithm is affected by several parameters, mainly the population size $I$, number of generations $N_G$ before termination, as  well as crossover $\rho_c$ and mutation $\rho_m$ rates. 
The crossover $\rho_c$ rate defines the probability of a crossover operation to generate a new offspring, while the mutation $\rho_m$ rate defines the probability of a random mutation of an individual in the population.
Both $\rho_c$ and $\rho_m$ determine the balance between exploration (discovering new areas of the search space), and exploitation (fine-tuning existing solutions).
In our actual practical use, moderate values were selected; namely for the crossover rate, we used $\rho_c=0.8$, and for the mutation rate $\rho_m=0.2$.
The population size $I$ and number of generations $N_G$ will be discussed in greater detail in the evaluation section.

\begin{figure*}[ht] % Use figure* for a figure spanning both columns 
\centering \includegraphics[trim={0cm 4cm 2cm 0cm},clip,width=0.85\textwidth]{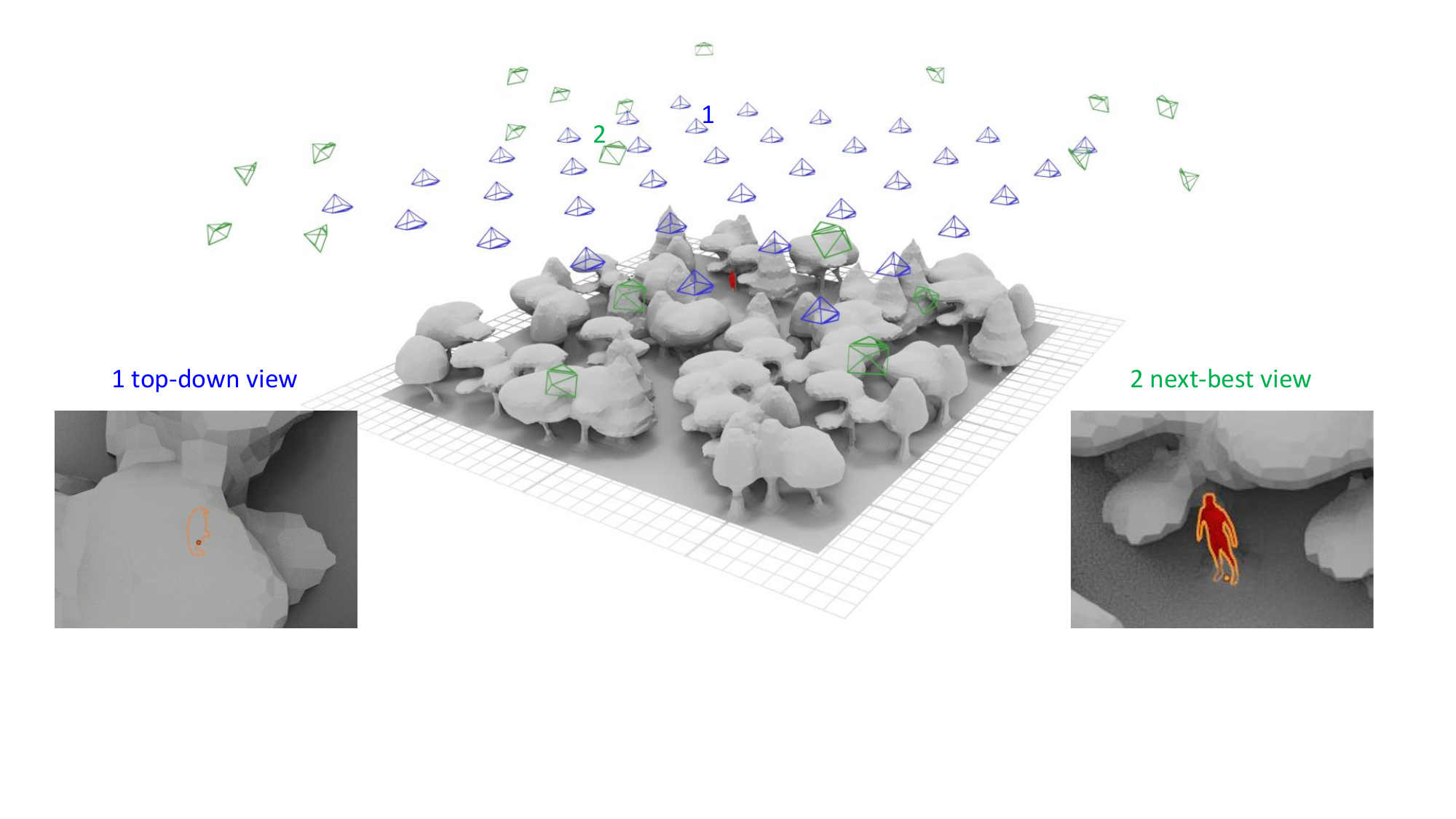} % Adjust the width parameter as needed 
\caption{Example of the outcome of a simulation run. The picture shows the randomized forest scene with one hidden manikin marked in red. The 36 blue cameras represent the initial views, while the 20 green cameras represent planned \ac{NBV} according to Visibility Fitness. On the left, a section of the top-down camera illustrates how the manikin is obstructed by the trees. In contrast, on the right, one of the planned camera views shows a clear visibility of the manikin.
} \label{fig:example} \end{figure*}

\section{Experiments and Evaluation}

To evaluate our approach we carried out experiments in simulations as well as in a real-world scenario.
The simulations allow us to get a deeper insight into the performance of our \ac{NBV} planning and to conduct a high number of experiments for different constellations of the environment.
On the other hand, our real-world experiments are less comprehensive but act as a proof of concept.
They show that our approach also works under real-world conditions.

%\todo{We want at least two different images to visualize the results. The first one is a graph that shows the number of detected object compared to the number of pictures/views.. Use these to compare the heuristics. A 3D object visualized with cameras pointing at it. One for each heuristics. Visually compare them.}

\subsection{Evaluation in Simulation} %references to the graphs need to be 

The purpose of the evaluation in simulation is twofold:
First, we want to verify that our approach can place cameras in a way so that the chosen heuristics are maximized and also converge to an optimum.
Second, we want to validate that by maximizing the heuristic we also increase our chance of finding hidden objects in the camera views. 
In addition, we also want to compare the two heuristics and which one performs better.
Let us start by explaining the simulation setup.

\begin{figure}[ht]
\centering
\centering
 \includegraphics[trim={0cm 0cm 0cm 0.0cm},clip,width=0.9\linewidth]{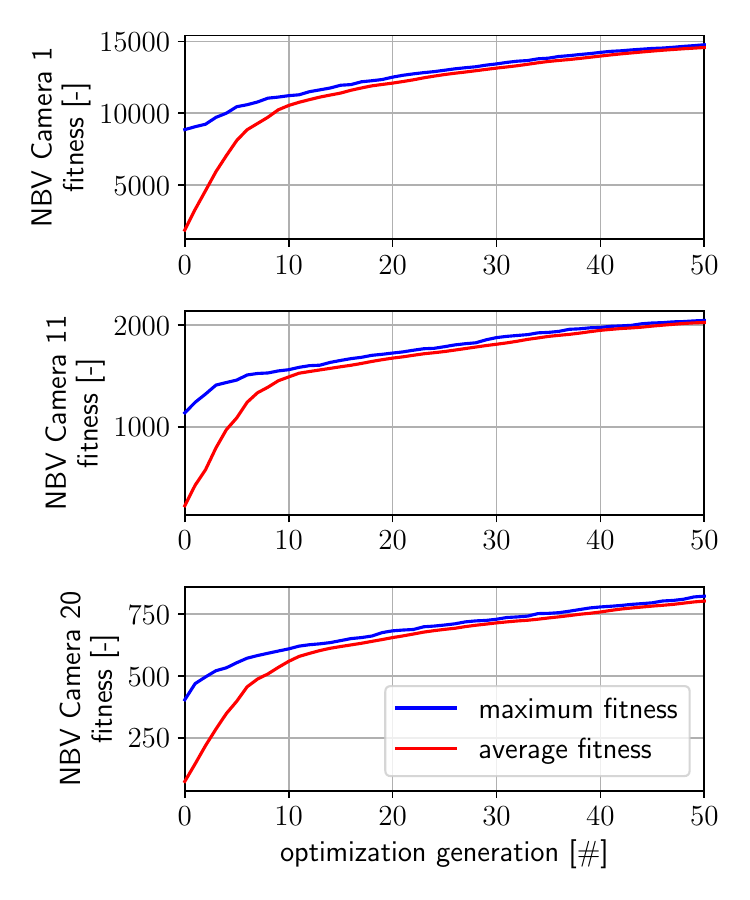}
\caption{Convergence of evolutionary algorithm for the $J_v$ fitness plotted over the number of generations. The blue curve represents the maximum achieved fitness in the population for a particular generation, while the red curve shows the average in the population.
}
\label{fig:saturation_d}
\end{figure}

\begin{figure}[ht]
\centering
\centering
 \includegraphics[trim={0cm 0cm 0cm 0.0cm},clip,width=0.9\linewidth]{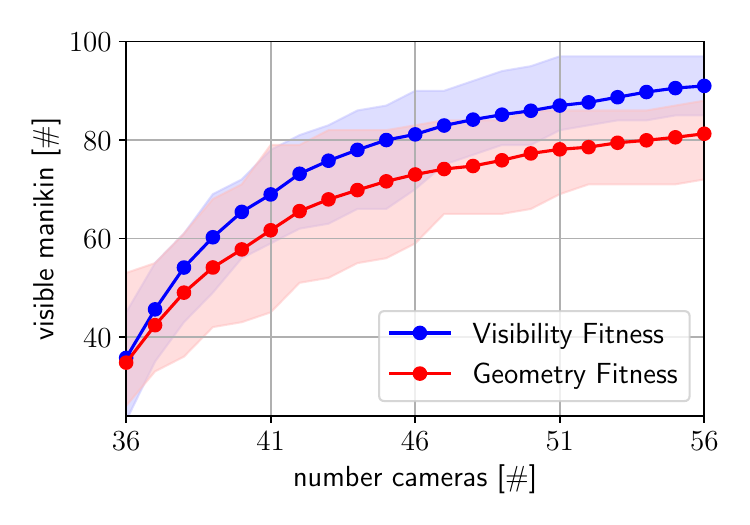}
\caption{The curves show how many of the 100 hidden manikins are visible with respect to the placed camera views. For the blue curve, the Visibility Fitness was used during the view optimization, and for the red curve the Geometry Fitness, respectively. The curves are averaged over 18 simulation runs with randomized environments. The transparent area marks the minimum and maximum values for the 18 runs.
}
\label{fig:visibility}
\end{figure}

\begin{figure}[ht]
\centering
\centering
 \includegraphics[trim={0cm 0cm 0cm 0.0cm},clip,width=0.9\linewidth]{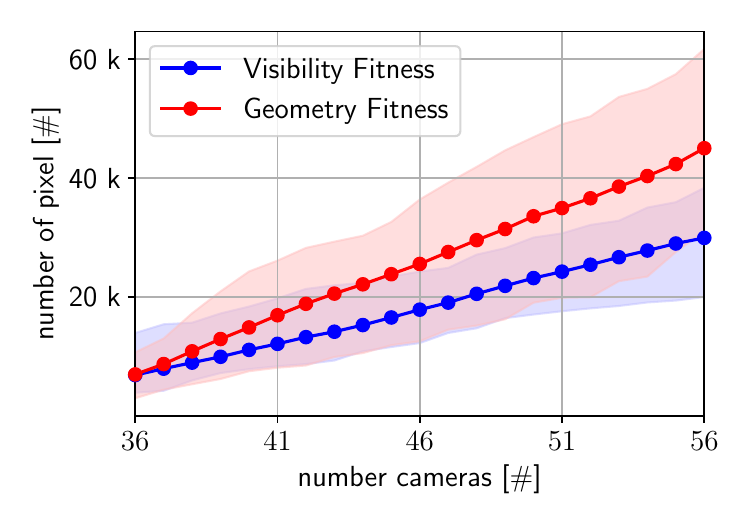}
\caption{The curves show how many red pixels (associated with the manikin) are seen when the scene is rendered for the camera views. The number of pixels is summed over all cameras and summed over all 100 hidden manikins. The curves are then averaged over 18 simulation runs with randomized environments. The transparent area marks the minimum and maximum values for the 18 runs.
}
\label{fig:pixel}
\end{figure}

\subsubsection{Simulation Setup}
The overall workflow of a single simulation run is depicted in \autoref{fig:workflow_sim}.
As mentioned in the \autoref{sec:ProblemDescription}, we assume that we start with a rough 3D model of the environment obtained from first surveillance from a top-down view of pictures taken by a \ac{UAV}.
In the simulation, we just generate a randomization forest scenario that spans 30 m $\times$ 30 m where trees are placed at random positions with random shapes and different densities.
Further, we place 36 initial cameras on a regular 6 $\times$ 6 grid above the scenario, which emulates the viewpoints from the first \ac{UAV} flight.
We are supposing that the 3D model has been created from these 36 camera views.
Based on this initial setup, we plan iteratively new camera views given the already placed cameras.
As a result, we obtain camera position $\vect{o}_{K+1}$, and orientation $(\theta_{K+1},\gamma_{K+1})$.
For the implementation, we make use of the Python library DEAP~\cite{DEAP_JMLR2012}.
In total, we plan 20 \acp{NBV}.
To evaluate the calculated \acp{NBV}, we hide 100 virtual red manikins in the forest environment below the trees at random positions.
For each camera, i.e. 36 initial cameras and 20 planned cameras, we render the scene once for each hidden manikin.
If the sum of red pixels in the rendered image is larger than 0, we consider the manikin visible by the camera view.
Based on that we can determine how many of the 100 manikins are at least partially visible by the placed cameras.
Where higher values mean better performance of the \ac{NBV} planing.
With this data, we can quantify the enhancement of visibility offered by the \ac{NBV} configuration when compared to the initial top-down perspectives.

\subsubsection{Simulation Results}
Let us first have a look at the performance of the \ac{EA} optimization.
In \autoref{fig:saturation_d}, the fitness value is plotted for finding the first, the 11th, and 20th \ac{NBV} after the initial 36 camera views.
The figure depicts the average and maximum fitness within one generation of the \ac{EA} optimization.
For these simulation runs, the population size was chosen as 50 and we made use of the Visibility Fitness, $J_v$.
The curves are averaged over 20 simulation runs with different randomized environments.
From the plots, a saturation effect can be observed after 20 generations.
For the Geometry Fitness, $J_d$, we observed a similar behavior.
This led us to terminate further simulation experiments after 20 generations since we consider more than twice as much optimization time not worth an increase of approximately $25\%$ in the heuristic value for our application.

The outcome of the \ac{NBV} planning is depicted in \autoref{fig:example}, as an example, with a hidden virtual manikin in red.
The scene shows the initial cameras with a top-down view in blue and the 20 planned camera views in green.
The manikin is not visible from the top-down views, since it is occluded by the trees.
However, one of the \acp{NBV} has a perfect view of the manikin.

\autoref{fig:visibility} shows how many of the 100 hidden manikins are visible in the placed camera views. 
The curves start with the initial 36 top-down cameras and show how the iteratively placed \acp{NBV} increase the chance of seeing manikins in the camera pictures.
The two curves utilize the two different heuristics in the \ac{EA} optimization of the views.
The plots are generated based on 18 simulation runs for each heuristic.
The initial views capture no more than a maximum of 55 and an average of 35 out of the 100 placed manikins. 
This indicates that approximately half of the manikins remain hidden from the initial views.
%As can be seen, the initial camera can only see less than 50$\%$ of the manikins.
By adding 20 cameras with our proposed \ac{NBV} approach based on the Visibility Fitness, we can see more than 90$\%$ of the manikins on average in the camera views.
For the Geometry Fitness, we increase the chance to 80$\%$ on average.

So from a comparison of the two heuristics, we can conclude that the Geometry Fitness heuristic does not enhance the ability to detect hidden individuals as much as the Visibility Fitness heuristic.
On average, the Visibility Fitness reveals 10$\%$ more individuals than the Geometry Fitness. 
However, it is noteworthy that the \ac{NBV}s calculated by the Geometry Fitness expose significantly more red pixels of hidden manikins when rendered, as can be seen from \autoref{fig:pixel}.
This suggests that while the Visibility Fitness is more effective in detecting at least some part of the manikins, the Geometry Fitness may offer a better view of the manikins at the price of missing some.

\begin{figure*}[ht] % Use figure* for a figure spanning both columns 
\centering \includegraphics[width=0.8\textwidth]{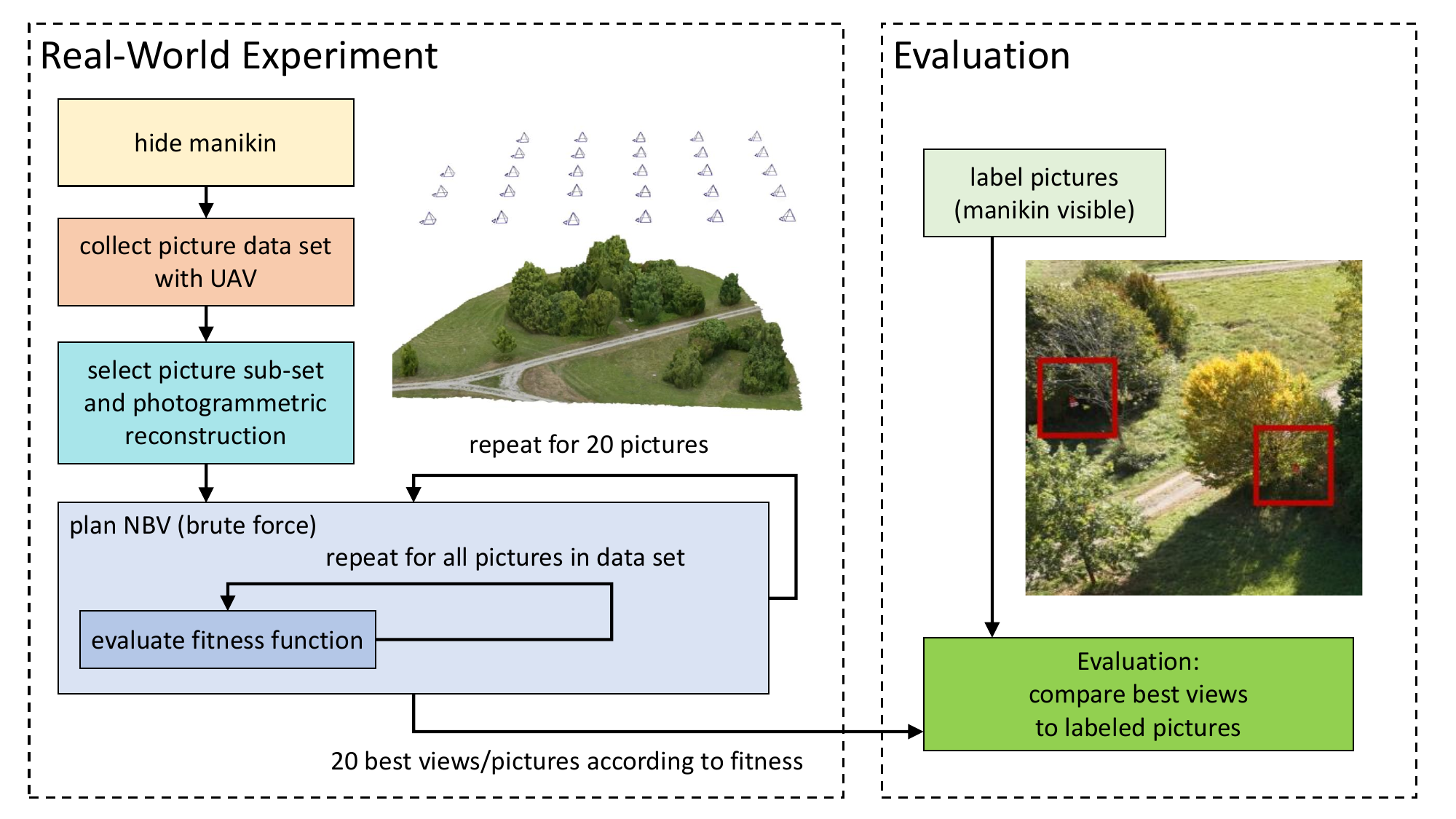} % Adjust the width parameter as needed 
\caption{Workflow of real-world experiments to evaluate the proposed \ac{NBV} planning.
} \label{fig:workflow_exp} \end{figure*}

\begin{figure}[ht]
    \centering
    \begin{subfigure}[b]{0.49\textwidth}
        \centering
        \includegraphics[trim={5cm 4cm 8.5cm 5.0cm},clip,width=0.99\textwidth]{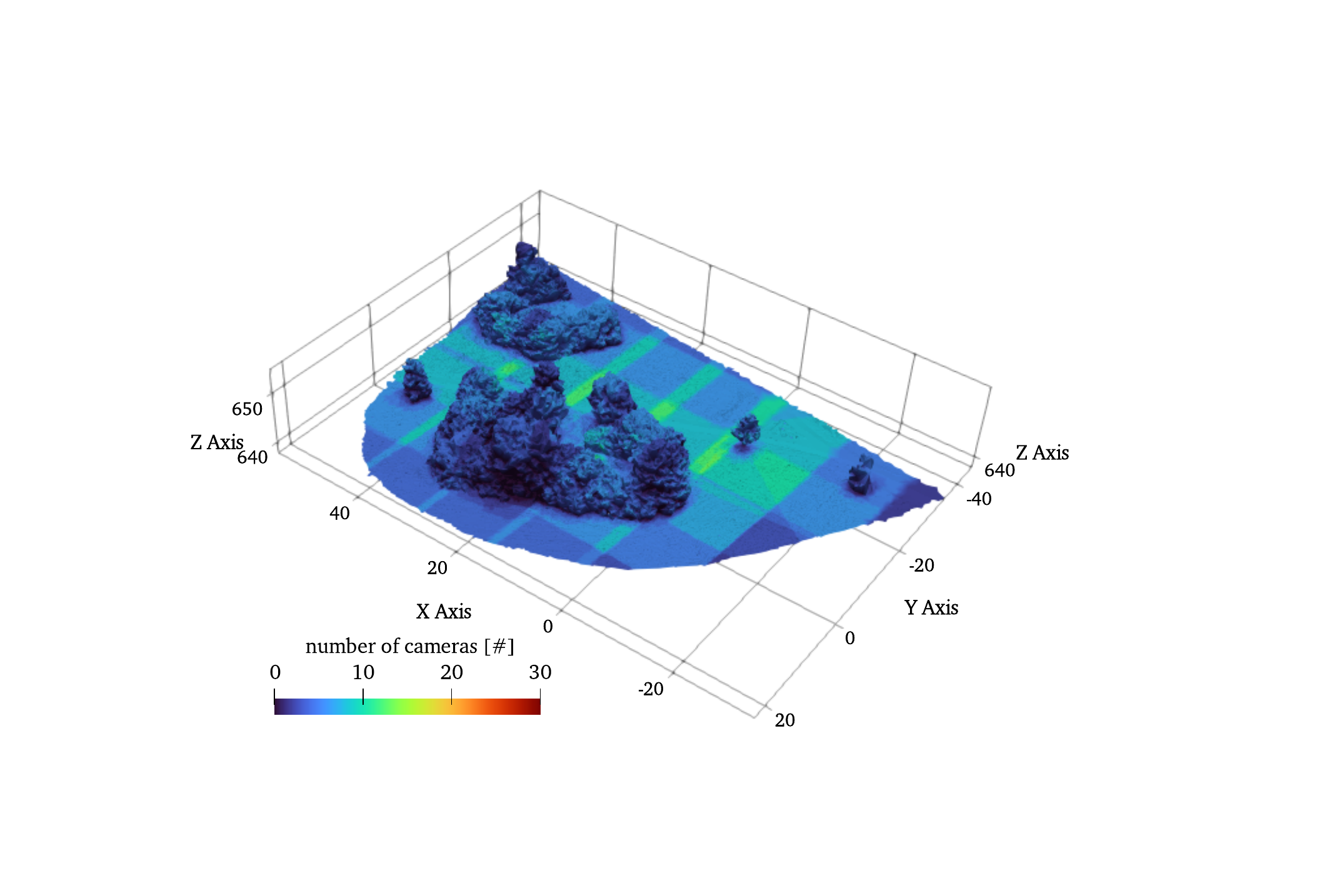}
        \caption{}
        \label{fig:initial}
    \end{subfigure}
    \hfill
    \begin{subfigure}[b]{0.49\textwidth}
        \centering
        \includegraphics[trim={5cm 4cm 8.5cm 5.0cm},clip,width=0.99\textwidth]{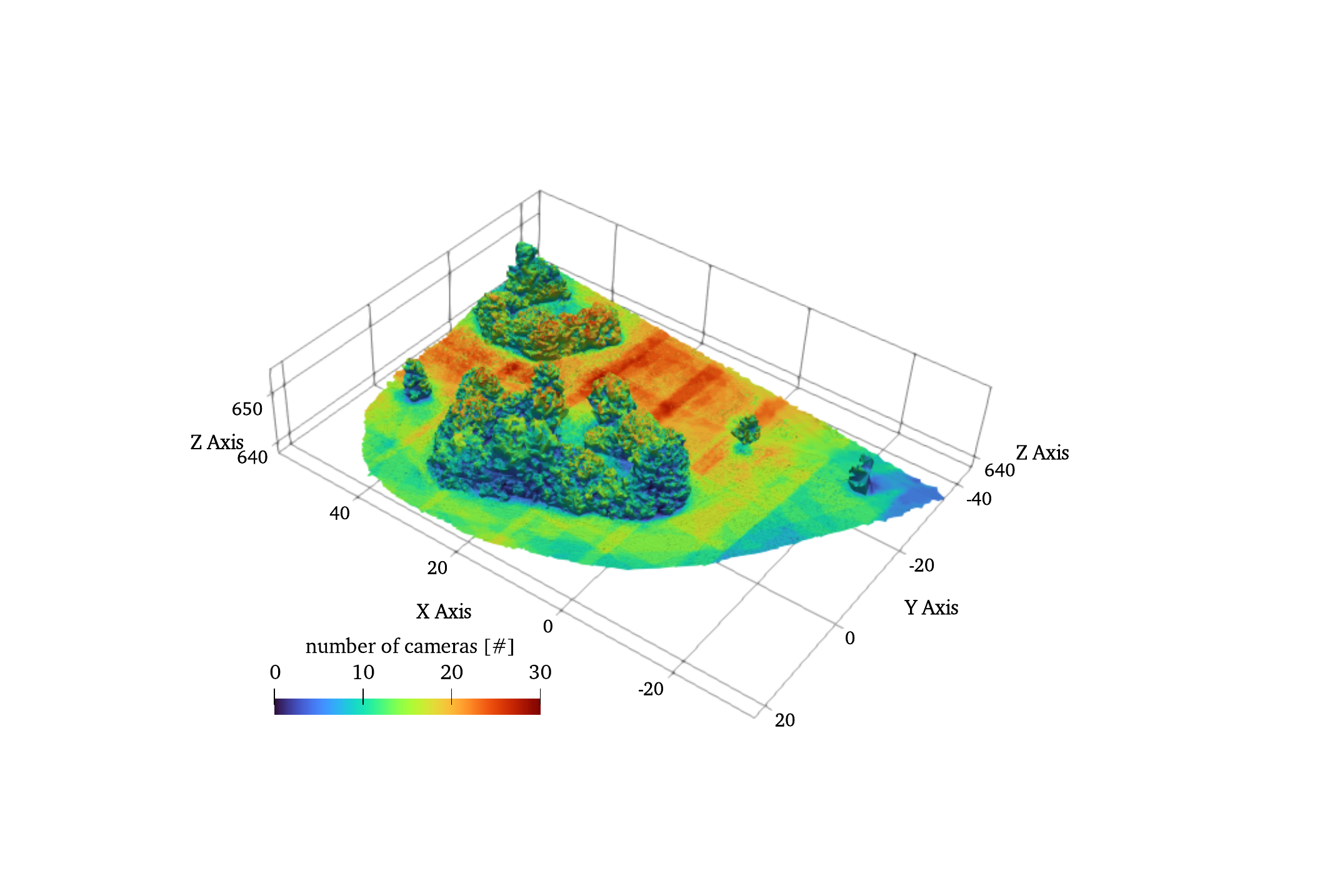}
        \caption{}
        \label{fig:nbv_v}
    \end{subfigure}
    \begin{subfigure}[b]{0.49\textwidth}
        \centering
        \includegraphics[trim={5cm 4cm 8.5cm 5.0cm},clip,width=0.99\textwidth]{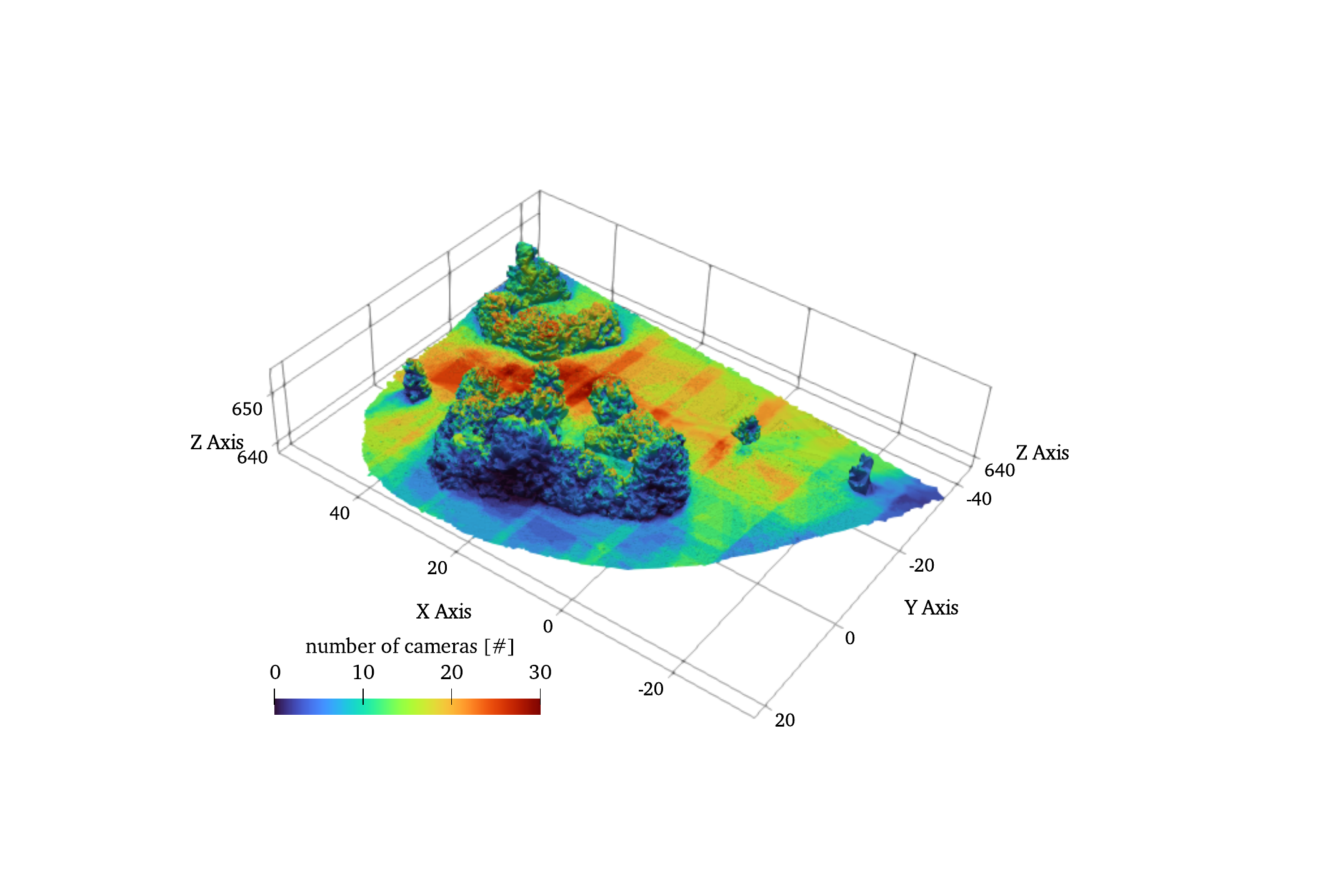}
        \caption{}
        \label{fig:nbv_d}
    \end{subfigure}
    \caption{Change in mesh vertex coverage after adding \acp{NBV} to the initial camera configuration. (blue: low coverage, red: high coverage). (a) shows the coverage for the initial camera configuration; (b) shows the coverage for 20 \acp{NBV} planned based on the Visibility Fitness and (c) for \acp{NBV} planned based on the Geometry Fitness. }
    \label{fig:configurations}
\end{figure}

\subsection{Evaluation in Real World Experiment}

%As mentioned above, we carried out real-world experiments as a proof of concept for our approach, in a simplified way compared to our more comprehensive simulation studies.

The overall workflow of the real-world experiments is depicted in \autoref{fig:workflow_exp}.
We have chosen a scenario with a group of bushes and some small trees surrounded by open space and a gravel street as shown in the 3D reconstruction in the middle of \autoref{fig:workflow_exp}.  
In this area, we hid manikins similar to the simulation.
For that, we placed seven tripods wearing high-visibility vests as markers to mimic our manikins.
These tripods were hidden near trees and in the bushes, making them difficult to detect by a top-down view, as shown in the picture on the right in \autoref{fig:workflow_exp} (red marked areas).
For the experiments, we collected a large data set of pictures taken from different positions above the scenario and with different yaw and pitch angles of the camera views.
To this end, a DJI Mavic 3 Pro was utilized.
The location and orientation of the pictures are saved as metadata for every picture.
In total, the data set covers 1032 pictures.
Further, every picture was manually labeled if a hidden manikin was visible. 
As a subset, we selected $31$ camera views on a grid pointing downwards as depicted in the middle of \autoref{fig:workflow_exp}.
This subset is considered our set of initial cameras, and we performed a photogrammetric reconstruction of the scenario based on this subset of pictures.
In contrast to the simulations, we do not run the \ac{EA} algorithm to find the \acp{NBV}.
Instead, we follow a brute-force approach and evaluate the fitness functions $J_v$ and $J_d$ for all pictures in the data set. 
The picture with the highest fitness score is selected as the \ac{NBV}.
Like in the simulation, we calculate 20 \acp{NBV} for our scenario.

As a result, 5 out of 7 placed manikins that were occluded in top-down views are repeatedly captured by the \ac{NBV} pictures highlighting the effectiveness of the approach also for practical applications.
\autoref{fig:configurations} illustrates the coverage of the 3D model, i.e. the number of cameras that can see a vertex of the 3D mesh.
Dark blue indicates no or only a few cameras can see an area, whereas red areas indicate a high visibility by many cameras.
\autoref{fig:initial} shows the coverage for the initial $31$ cameras.
Particularly near the ground beneath the trees, the coverage is very limited.
In \autoref{fig:nbv_v} the coverage is shown in case the pictures are scored according to the Visibility Fitness $J_v$.
The result shows a significantly higher coverage than for initial views, with less blue areas below trees and more orange in open fields.
Lastly, the coverage using the Geometry Fitness $J_d$ is shown in \autoref{fig:nbv_d}.
The plot shows many more red spots compared to \autoref{fig:nbv_v} indicating a higher peak coverage.
However, this improvement is mainly limited to open field areas.
At the same time, areas below trees show worse coverage compared to \autoref{fig:nbv_v} and $J_v$. Thus, it can be concluded that $J_v$ demonstrates a more consistent improvement in coverage through camera views compared to $J_d$.

\section{Conclusions}

In this work, we focused on the \ac{NBV} problem for selecting strategic viewing positions for \ac{UAV}s, increasing the likelihood of locating missing persons in \ac{SAR} operations, where visibility is obstructed.
We developed and tested two novel heuristics for optimizing cameras' position and orientation in both real-world and simulated applications. We demonstrated that these heuristics enhance \ac{UAV}-assisted searches and can be seamlessly integrated into \ac{SAR} planning. The Visibility Fitness heuristic, $J_v$, has shown greater scene coverage, identifying over 90\% of the hidden manikins on average in simulations, compared to 80\% detected by the Geometry Fitness function, $J_d$. In real-world experiments, $J_v$ achieved better area coverage beneath the canopy.

Despite these advancements, current limitations in the performance of \ac{NBV} calculations remain a significant challenge. As the number of required camera views increases, so does the computational time needed to calculate them. To address this, future work will focus on optimizing \ac{NBV} calculations and improving solution quality. Furthermore, \ac{UAV}s will be autonomously tested in search and rescue missions in dense forest environments and disaster-stricken areas, highlighting their potential for overcoming the challenges of obstructed and inaccessible terrains.

% use section* for acknowledgment
\section*{Acknowledgement}
This research was funded by internal German Aerospace Center (DLR) funds. The authors declare that they have no competing interests.

\begin{acronym}
\acro{NBV}{Next Best View}
\acro{UAV}{Unmanned Aerial Vehicle}
\acro{RRT}{Rapidly-exploring Random Tree}
\acro{SAR}{Search and Rescue}
\acro{EA}{Evolutionary Algorithm}
\acro{UGV}{Unmanned Ground Vehicle}
\acro{RGB}{Red Green Blue}
\acro{LiDAR}{Light Detection and Ranging}
\end{acronym}

% Can use something like this to put references on a page
% by themselves when using endfloat and the captionsoff option.
\ifCLASSOPTIONcaptionsoff
  \newpage
\fi

\printbibliography

\end{document}